\documentclass[letterpaper, 10 pt, conference]{ieeeconf}  

\IEEEoverridecommandlockouts                              
\usepackage [table]{xcolor}
\usepackage[cmex10]{amsmath}

\usepackage[nodisplayskipstretch]{setspace}

\usepackage{graphics} 
\usepackage{epsfig} 
\usepackage{epstopdf}
\usepackage{subfig}
\usepackage{mathptmx} 
\usepackage{times} 
\usepackage{amsmath} 
\usepackage{amssymb}  
\usepackage{algorithm}
\usepackage{algpseudocode}
\usepackage{array}
\usepackage{bm}
\usepackage{cite}
\usepackage{xcolor}
\usepackage{float}
\usepackage{multirow}
\usepackage[font=small]{caption}
\usepackage{siunitx}
\usepackage{caption}
\usepackage{ctable}
\usepackage{array}
\usepackage{tensor}

\newcolumntype{P}[1]{>{\centering\arraybackslash}p{#1}}

\title{A Vision-Guided Multi-Robot Cooperation Framework for Learning-by-Demonstration and Task Reproduction}

\author{Bidan Huang,
        Menglong Ye,
        Su-Lin Lee,
        Guang-Zhong~Yang,~\IEEEmembership{Fellow,~IEEE}
\thanks{B. Huang, M. Ye, S.-L. Lee and G.-Z. Yang are with the Hamlyn Centre for Robotic Surgery, Imperial College London, SW7 2AZ, London, UK (e-mail: b.huang@imperial.ac.uk). The project is supported by the EPSRC (EP/L020688/1).
}}

\begin{document}

\maketitle
\thispagestyle{empty}
\pagestyle{empty}

\begin{abstract}
This paper presents a vision-based learning-by-demonstration approach to enable robots to learn and complete a manipulation task cooperatively. With this method, a vision system is involved in both the task demonstration and reproduction stages. An expert first demonstrates how to use tools to perform a task, while the tool motion is observed using a vision system. The demonstrations are then encoded using a statistical model to generate a reference motion trajectory. Equipped with the same tools and the learned model, the robot is guided by vision to reproduce the task. The task performance was evaluated in terms of both accuracy and speed. However, simply increasing the robot's speed could decrease the reproduction accuracy. To this end, a dual-rate Kalman filter is employed to compensate for latency between the robot and vision system. More importantly, the sampling rates of the reference trajectory and the robot speed are optimised adaptively according to the learned motion model. We demonstrate the effectiveness of our approach by performing two tasks: a trajectory reproduction task and a bimanual sewing task. We show that using our vision-based approach, the robots can conduct effective learning by demonstrations and perform accurate and fast task reproduction. The proposed approach is generalisable to other manipulation tasks, where bimanual or multi-robot cooperation is required.  

\end{abstract}


\section{Introduction}

Learning-by-demonstration has become one of the standard approaches for performing a range of tasks~\cite{pan2012recent}. Increasing numbers of industrial robots now have built-in ``record-and-replay'' modes, which allow users to physically move their joints and then replay the same trajectory. This function enables workers without robotic expertise to program a robot trajectory and complete simple tasks; however, there are limitations to this kinesthetic teaching approach. Firstly, it requires a robot with this ``record-and-replay'' mode during the demonstration, which might not always be available. Secondly, the user needs to kinesthetically move the robot arm to demonstrate a task: for a small and delicate manipulation task such as sewing or surgical subtasks~\cite{osa2014trajectory,murali2015learning,sen2016automating}, this can be difficult as these tasks are usually demonstrated via teleoperation or use pre-programmed trajectories. Thirdly, it is difficult to accurately demonstrate a bimanual task by kinesthetic teaching; typically a user must use both hands to move a robot arm. In industry, bimanual tasks are usually demonstrated by moving the robot with the control panel or with pre-programmed commands~\cite{makris2014intuitive}.

\begin{figure}
\centering
\includegraphics[width=8cm]{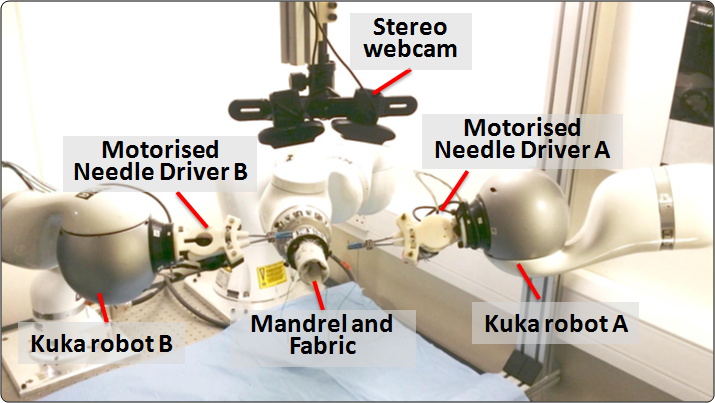}
\caption{An overview of our system setup. Two Kuka robots hold the motorised needle drivers to perform bimanual sewing on a target under the observation of a stereo camera. An addition robot is used to hold the target (mandrel and fabric) for manipulation. }
\label{fig:setup}
\end{figure}

Transferring skills from users to robots can be performed naturally via vision-based demonstration ~\cite{bandera2012survey}. With a vision system observing the user motion, no robot is required during the demonstration and the user can perform the task naturally with their own hands, e.g. manipulating an object or using a tool. This type of vision-based demonstration learning facilitates the performance of delicate tasks with high precision. Furthermore, bimanual or multiple tool/robot cooperation tasks require accurate tool/robot co-localisation. At the demonstration stage, a vision system can provide the configurations of multiple tools/robots and objects in a unified coordinate frame, via tracking their poses in real-time. This facilitates finding the correlation between tools and objects in learning. At the task reproduction stage, the poses of the robots and objects can be similarly retrieved via tracking. This then enables the robots to perform accurate task execution via visual servoing, which mitigates the need for accurate hand-eye calibration.
Despite the benefits of a vision-based learning-by-demonstration approach, two main problems should be addressed: how to map the user's behaviour to the robot behaviour and how to achieve good task performance, e.g. high accuracy and speed.

To tackle these problems, this paper proposes a vision-based learning by demonstration approach for task manipulation, which allows easy manipulation learning and accurate task execution. To learn a manipulation task, we adopt an ``object centric'' approach~\cite{phung2012tool,li2014learning}. Instead of modelling the human hand or robot motion, we model the motion of manipulated objects during demonstration learning. With this, we do not need to consider a direct mapping from the human joints motion to the robot joints motion, whilst we focus only on reproducing the same object behaviour. In our approach, we assume that the target object is rigidly attached to the robot end-effector and hence its behaviour can be reproduced by transforming its motion to the end-effector.


As the demonstrations are observed and learned in the coordinate frame of the vision system, programming the robot to perform the tasks under the same frame with visual guidance is an intuitive solution. We use the ``look-and-move’’ visual servoing approach~\cite{hutchinson1996tutorial} to instruct the robot to complete the learned task. For reliable visual servoing, vision-based detection and tracking has been studied in the past decade for industrial~\cite{Lippiello2007,Cai2013} and surgical robotic setups~\cite{Mebarki2010,Zhang2016}. Accurate pose estimation of tools during manipulation plays an important role in guiding robots to follow desired trajectories for its provided advantage of compensating hand-eye calibration or kinesthetic errors~\cite{Ye2016}.

A robust vision-based method is proposed to guide the robot to perform reliable task execution. A dual-rate Kalman filter~\cite{wang2015statistical} is applied to deal with the low sampling rate and the latency between vision and kinesthetic data. With accurate estimation of the visual feedback, robot can perform the task accurately. To further overcome the slow visual feedback and increase robot speed, we optimise the reference trajectory based current task context, and adjust the robot speed accordingly.

Here, learning-by-demonstration is combined with visual servoing. The main contributions of this work are:

\begin{enumerate}

\item An easy-to-use method is introduced for users to demonstrate delicate bimanual tasks by simply using their own hands without the need for using robots. This allows for more natural demonstrations without requiring the user to directly adapt complex motions for the robots.

\item A robust visual servoing approach is proposed, which uses low-cost cameras to guide robots for task reproduction. The approach enables robot motion speed to be adjusted according to the task context, such that the speed of reproduction can be increased without sacrificing the reproduction accuracy.

\end{enumerate}

The rest of the paper is organized as follows. Section II describes our system, the hardware, and the software components. Section III shows the experiments conducted using this system and presents the results. Discussion and conclusions are presented in Section IV.

\begin{figure}
\centering
\includegraphics[width=6cm]{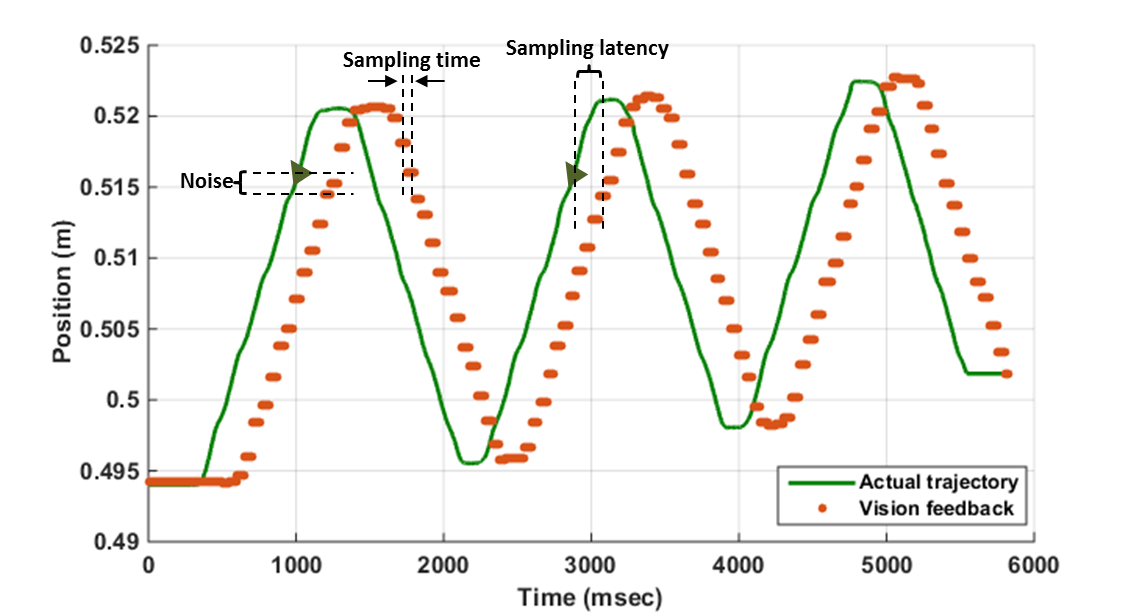}
\caption{An illustration of vision sensing latency and noise.}
\vspace{-0.6cm}
\label{fig:latency}
\end{figure}

\section{System Architecture}
The main motivation of this work is to present a system to generically solve the problem of a robot learning a bimanual task via vision. We focus on tasks of using tools and manipulating objects. Here, a ``tool’’ is a device that can be either held by a user or attached rigidly to a robot end-effector, whilst an ``object’’ can be manipulated in 3D space by tools. The user and robot use the same tools and objects during demonstration and reproduction. Learning is done using the object centric principle. After modelling the manipulated objects’ or tools’ motion, these models are applied to robots to reproduce the behaviour. Section A details our method of tracking tools and the process of user demonstration with a given bimanual task. Section B shows the encoding of the demonstrated motion and the optimization of the reference trajectory according to task contexts. Section C explains our implementation of the Kalman filter for the vision feedback control.

Without loss of generality, our approach is framed in the context of a bimanual sewing task for the purpose of personalized stent graft manufacturing~\cite{huang2016vision}. This task is challenging as it involves complicated movements and fine manipulation of the needle and requires high reproduction accuracy, e.g. pierce in and out of fabric within a 2mm slot, passing the needle from one robot to another. In our previous work, the robot was programmed by kinesthetically teaching one arm to complete the entire stitch cycle. This system has an uncertainty: the needle could change its pose when detached from the needle driver by unpredictable fabric tension. This introduces a large amount of uncertainty in the sewing process. The bimanual sewing system presented in this paper is for the purpose of reducing the uncertainty and increase the robustness of the process.

In this work, the vision system is extended to learn bimanual sewing. There are three different devices: a mandrel\footnote{A mandrel is a hollow cylinder for bounding the fabric and the stents together. It is design according to patients' anatomy. Details about it please see our previous work~\cite{huang2016vision}} for fixing the fabric, a needle for sewing, and needle drivers for gripping the needle. The proposed platform is shown in Figure~\ref{fig:setup}: one robot arm holds the mandrel and fabric while two robot arms with needle drivers manipulate a surgical needle. The needle drivers control the needle to pierce the fabric and complete a stitch. We focus on learning the trajectories of the needle drivers (tools) and the needle (object).

\subsection{Vision-based Bimanual Task Demonstration}
\label{sec:demonstration}
For both the task demonstration and reproduction, the same stereo vision system was used to avoid any error caused by different vision systems. The vision system plays two roles: motion observation during task demonstration and robot servoing for task reproduction.

\subsubsection{Detection-Tracking for Continuous Tool Pose Estimation}
\label{sec:vision}

Our stereo vision system was composed of two cameras calibrated using the OpenCV~\footnote{http://opencv.org/} library. To facilitate pose estimation, several fiducial bar-code markers were placed on each tool such that the entire motion trajectory can be observed. As shown in Figure \ref{fig:markers}, a hollow pentagonal adapter was designed and attached rigidly to each needle driver, with each adapter face being a 1cm$^{2}$ square on which the bar-code maker was placed. Each marker had a different pattern so that the rotation of the needle driver can be computed correctly. The detected marker pose was transformed to the needle driver pose according to the pentagonal prism's size. The rotational axis of the pentagonal prism was designed to align with the needle driver. For the mandrel, an octagonal prism was designed.

To allow for continuous 3D pose estimation of the tools in consecutive frames, a vision-based marker detection-tracking approach was therefore adopted. The marker detector was an off-the-shelf approach provided in ArUco~\cite{Jurado2014} which processes only the current frame, whilst the tracker was a forward-backward (FB) approach \cite{Kalal2010} based on optical flow that considers temporal information.

More specifically, the tracker was applied on a marker whose previous pose was available but whose current pose could not be detected in the frame. With previous locations $\left\lbrace q_{i}\right\rbrace_{i=1}^{n}$ of the corner points on a marker, optical flow was applied to forwards track from the previous to the current frame, to obtain the corner estimates $\left\lbrace q^{+}_{i}\right\rbrace_{i=1}^{n}$. Then, optical flow was applied again to backwards track these corner estimates from the current to the previous frame, thus obtaining additional estimates as $\left\lbrace q^{-}_{i}\right\rbrace_{i=1}^{n}$. With these, to determine if a corner point estimate $q^{+}_{i}$ was valid, its FB error, defined as Euclidean distance $\epsilon\left(q_{i},1^{-}_{i} \right)$, was compared to $\tau$. The value of $\tau$ was chosen as 1 pixel in this work, which helps filtering out the outlier corner estimates. The remaining estimates were treated as inliers, which were then used to estimate the 6 d.o.f pose of the marker via perspective-n-points~\cite{Lepetit2009}. Note that, when $\tau$ was larger than $5$ pixels, the output was empirically ignored from the tracker, as it indicated the corner estimates were not reliable.

Therefore, this tracker was able to provide pose estimation when the detector failed, thus providing continuous pose estimation. This detection-tracking strategy was applied to every marker that was attached to the tools.

\subsubsection{Data Acquisition}

In the bimanual sewing task, the user held a needle driver in each hand to manipulate the needle. The fabric was bound to the mandrel to constrain its deformation. The stereo camera was placed at the top of the mandrel so that the sewing behaviour, i.e. the needle drivers and needle motion, could be observed. The 6 d.o.f poses of both needle drivers were tracked and recorded by the stereo camera.

\begin{figure*}
\centering
\includegraphics[width=16cm]{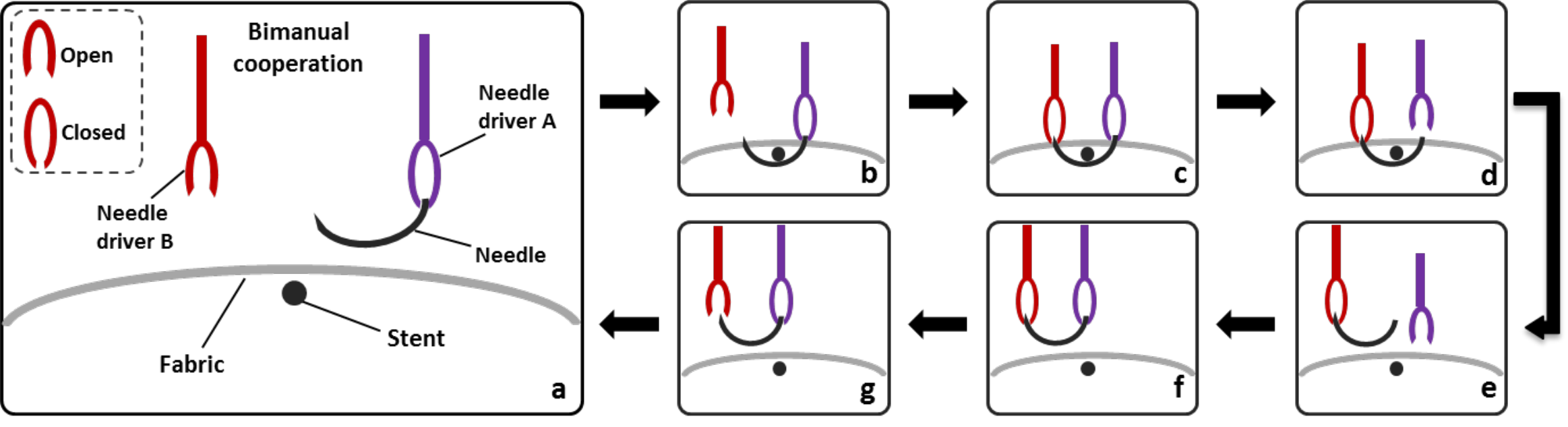}
\caption{The key steps involved in one stitch cycle (a-g), of which at the end the needle is passed back to needle driver A for next cycle.}
\label{fig:stitchcycle}
\end{figure*}

The motion of finishing one stitch is treated as one cycle; the steps for this are shown in Figure \ref{fig:stitchcycle}. The two needle drivers are referred to as ``Needle Driver A'' and ``Needle Driver B''. A single tip surgical curved needle was used. The sharp end of the needle is the ``needle tip'' and the blunt end is the ``needle end''. While Needle Driver B was static in our previous work, here both needle drivers were mobile. At the beginning of the task, the curved needle was gripped by Needle Driver A at its end (Step a). Needle Driver A approached the fabric and pierced it with the needle (Step b). When the needle tip pierced out of the fabric, Needle Driver B approached the needle tip and gripped it (Step c). Needle Driver A then opened and released the needle (Step d). After that, Needle Driver B pulled the needle out of the fabric (Step e). When the needle was completely out, Needle Driver A approached it and re-gripped it from the needle end (Step f). Needle Driver B then opened and released the needle (Step f). Finally Needle Driver A pulled the thread to tighten the stitch and returned to its initial pose. Hence a full stitch cycle, i.e. one stitch, was completed.

In this bimanual task, the two needle drivers hold the needle alternately. When a needle driver was holding the needle, the needle trajectory was recorded in the world frame; when the needle driver was not holding the needle, its own trajectory was recorded in the local frame of the needle. The needle trajectory was not directly observed but was computed as a rigid transformation of the observed needle driver position; this transformation was estimated by a needle pose detection method~\cite{huang2016vision}. This detection was performed at the beginning of each stitching cycle.

\subsection{Task Learning}
After multiple user demonstrations, e.g. five times for bimanual sewing, the different demonstration data were temporally aligned using Dynamic Time Warping ~\cite{berndt1994using}. According to the needle drivers' open and closed status and their attachment to the needle, one stitch cycle was segmented into four motion primitives, as shown in Table~\ref{tab:segments}. Each segment was a primitive movement and encoded by a Gaussian Mixture Model (GMM) for its ability to encode non-linear data~\cite{calinon2007learning,Huang2013}.

\begin{table}
\fontsize{8}{12}\selectfont
\centering
\caption{Motion primitives of stitching}
\vspace{5pt}
\label{tab:segments}
\hspace{-0.5cm}
\begin{tabular}{P{1cm}P{0.8cm}P{1.5cm}P{1.5cm}P{1cm}}
\hline
\rowcolor[HTML]{C0C0C0}
\begin{tabular}[c]{@{}c@{}}Motion\\Primitives\end{tabular} & \begin{tabular}[c]{@{}c@{}}Steps in \\ Fig. 3\end{tabular} & \begin{tabular}[c]{@{}c@{}}Needle Driver\\A status\end{tabular} & \begin{tabular}[c]{@{}c@{}}Needle Driver\\B status\end{tabular} & \begin{tabular}[c]{@{}c@{}}Needle\\status\end{tabular} \\ \hline \hline
1. & a, b & Closed & Open & With A \\ \hline
2. & c & Closed & Closed & With A  \\ \hline
3. & d,e & Open & Closed & With B \\ \hline
4. & f & Closed & Closed & With B \\ \hline
5. & g & Closed & Open & With A \\ \hline
\end{tabular}
\vspace{-0.2cm}
\end{table}

In this study, each motion primitive was modelled by a 7 d.o.f GMM $\Omega$, which encoded the time stamp $t$ and the 6 d.o.f pose $h = \{x,y,z,\alpha,\beta,\theta\}$. The probability that a given data point ${t,h}$ belonged to $\Omega$ was computed as:

\begin{equation}
\setstretch{1.5}
p\left(t,h\mid\Omega\right) = \sum_{m=1}^M \pi_m p_m\left(t,h\mid\mu_m,\Sigma_m\right)
\end{equation}
where $\pi_m$, $p_m$, ${\mu}_m$, and ${\Sigma}_m$ were the prior, the corresponding conditional probability density, mean, and covariance of the $m$-th Gaussian component, respectively. The number of Gaussian components was $M$ and was determined by a five-fold cross validation.

%

%
%

For each time step, Gaussian Mixture Regression (GMR) was used to generate trajectory points and hence form a reference trajectory to the demonstrated task.

\subsubsection*{Trajectory Optimisation for Task Contexts}
The learned reference trajectory must be further optimised to allow the robots to complete the task with high speed and high accuracy. In a manipulation task, there are two task contexts: ``end point driven'' and ``contact driven''. In the ``end point driven'' context, the robot can be moved freely as long as it reaches the final destination. In the ``contact driven'' context, the robot is in contact with the environment and its motion is constrained; in this context, the robot movement needs to follow a particular trajectory.

In the case of bimanual sewing, the approaching motion of the needle to the fabric is end point driven, while the piercing in and out motion is contact driven. For the approaching motion, the needle does not need to follow the reference trajectory from point to point and hence the trajectory can be down-sampled to allow the robot to move faster. For the piercing motion, the needle needs to follow the reference trajectory and accuracy is the first priority. Hence the robot must slow down and follow the reference trajectory carefully.

The task context was distinguished by the variance between different demonstrations. Figure ~\ref{fig:demo} shows the variance of  Motion Primitive 1 across the five demonstrations. The variance of the trajectories is large at the beginning, i.e. it is end point driven, and low at the end, i.e. it is contact driven, hence the beginning of this trajectory was down-sampled. The correlation of the variance and the ratio to the demonstration speed $r$ was:

\[
    r =
\begin{cases}
 0.5,& \text{if } v_t > 0.01 m \text{ $or$ } v_r > 0.2 rad\\
 1 & \text{if }  0.01 m > v_t > 0.005 m \text{ $or$ } \\
   & 0.2 rad > v_r > 0.1 rad\\
 2 & \text{if } v_t<0.005 m \text{ $or$ } v_r<0.1 rad \\
\end{cases}
\]
where $v_t$ and $v_r$ are the variance of the translation and the rotation, respectively. This is decided according to our system and the task requirement (Session~\ref{sec:experiment}).



\subsection{Task Reproduction with Vision Guidance}
\label{sec:visualservoing}
\begin{figure}
\centering
{
\includegraphics[width=6cm]{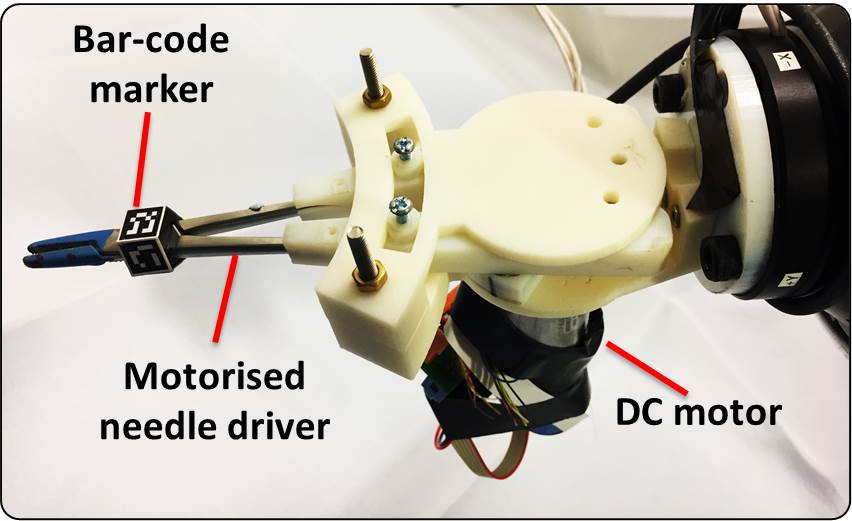}
\caption{The motorised needle driver. It is designed for attachment to the Kuka robot and has a DC motor that opens and closes the needle driver.}
\label{fig:needledriver}
}
\vspace{-0.6cm}
\end{figure}

To allow the robot to reproduce the task, the tool was first motorized and installed on the robot. Figure~\ref{fig:needledriver} shows the motorized needle driver designed for the robot. The visual markers were placed at the same location, such that the task can be reproduced by following the reference trajectory.

\subsubsection{Visual Servoing}
For robots with considerable kinematic errors, online vision feedback is essential. For high accuracy industrial robots, online vision feedback is also important, especially for multi-robot operation or bimanual tasks where all robots need to be registered to the same frame. For delicate tasks such as sewing or surgical tasks, it is very time-consuming to achieve a good off-line calibration with adequate precision. Therefore our robots were controlled with online vision feedback, such that the error of the robot reaching the target is independent of the registration and the robot kinematic accuracy~\cite{hutchinson1996tutorial}.

All the robots in this study were registered to the camera frame. This was performed by computing the transformation between the robot ($r$) base and the camera ($c$) base $\tensor[^c]{H}{_r}$ according to the pose of the markers relative to the robot base frame and the camera frame\footnote{We denote $\tensor[^b]{H}{_a}$ as the homogeneous matrix of the pose of an object a in the frame of an object b.}. In the bimanual sewing task, the needle driver ($d$) pose was first computed in the robot end effector ($EE$) frame $\tensor[^{EE}]{x}{_d}$ according to the motorized needle driver design. The needle driver was then moved into the view of the camera and a series of end effector poses $\tensor[^r]x{_{EE}}$, as well as the needle driver pose under the camera frame $\tensor[^c]x{_d}$, was recorded at each time step. The transformation between the robot base frame and the camera frame $\tensor[^c]{H}{_r}$ was then estimated by the absolute orientation algorithm \footnote{From Matlab: https://uk.mathworks.com/matlabcentral/fileexchange/22422-absolute-orientation}.


The ``look-and-move'' position based servoing approach was used for our tasks. In this method, the problem is modelled as moving the robot to reduce the error between the current pose of the manipulated object and its target pose. Taking Motion Primitive 1 in our sewing task as an example, the aim is to move the needle to approach the mandrel and pierce the fabric. The reference trajectory was expressed in the frame of the mandrel ($m$) as a series of target poses of the needle ($n$) $\tensor[^m]x{_{n^*}}$. These needle target poses were transferred to needle driver ($d$) target poses $\tensor[^m]x{_{d^*}}$ :

\begin{equation}
\setstretch{1.5}
{
\tensor[^m]x{_d^*} =
\tensor[^m]x{_{n^*}}\cdot
\left(\tensor[^d]H{_n}\right)^{-1}
}
\end{equation}
where $\tensor[^d]H{_n}$ is the needle and needle driver relative pose detected over the task as explained in Section~\ref{sec:demonstration}.

As the needle driver and the mandrel was directly observed by the camera ($c$), the error of the pose was computed as

\begin{equation}
\setstretch{1.5}
{
\tensor[^d]x{_{d^*}} =
\left(\tensor[^c]x{_{d}}\right)^{-1}\cdot
\tensor[^c]x{_m}\cdot\tensor[^m]x{_{d^*}}
}
\end{equation}

This then can be transformed to the error of the robot end effector and hence generate commands to move towards the target pose.

\begin{figure}
\centering
{
\includegraphics[width=6cm]{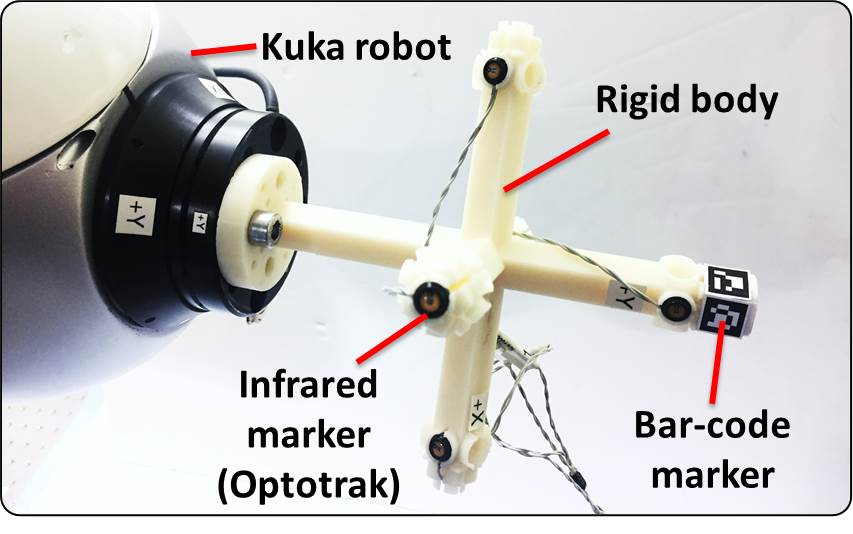}
\caption{A 3D printed rigid body for validation, with the Optotrak Certus infrared markers and the bar-code markers affixed.}
\vspace{-0.6cm}
\label{fig:markers}}
\end{figure}

\subsubsection{Kalman Filter}
In relative terms, camera latency is inevitable compared to other sensors used in robotic control.  To balance these, a dual rate Kalman filter was used to estimate the vision feedback $\tensor[^c]x{_{d}}$.

In a standard Kalman filter, the state of a system at a time $t$ predicted from the previous state at time $t-1$ is modelled as

\begin{equation}
\setstretch{1.5}
\boldsymbol{x}_t = \boldsymbol{F}_t\boldsymbol{x}_{t-1} + \boldsymbol{B}_t\boldsymbol{u}_t+\boldsymbol{w}_t
\end{equation}
where $\boldsymbol{x}_t$, $\boldsymbol{F}_t$, $\boldsymbol{B}_t$, $\boldsymbol{u}_t$, and $\boldsymbol{w}_t$ are the state vector, the state transition matrix, the control input matrix, the control inputs, and the process noise term at time $t$, respectively. The noise term is assumed to be a zero mean Gaussian with covariance matrix $\boldsymbol{Q}_t$.

The system measurement is modelled as

\begin{equation}
\setstretch{1.5}
\boldsymbol{z}_t = \boldsymbol{H}_t\boldsymbol{x}_t + \boldsymbol{v}_t
\end{equation}
where $\boldsymbol{z}_t$, $\boldsymbol{H}_t$, and $\boldsymbol{v}_t$ are the measurement vector, the transformation matrix which transforms the system state space to the measurement space, and the measurement noise at time $t$, respectively. The noise term is also assumed to be a zero mean Gaussian with covariance matrix $\boldsymbol{R}_t$.

In our sewing task, both the system state and measurement are the 6 d.o.f needle driver pose. Hence the transformation matrix $\boldsymbol{H}_t$ is an identity matrix. Similarly, the state transition matrix $\boldsymbol{F}_t$ is also an identity matrix. These are omitted in the following equations. The Kalman filter has two phases: prediction and measurement update. Applying the Kalman filter to our task, the equations of the prediction phase are

\begin{equation}
\label{equ:predict}
\setstretch{1.5}
\hat{\boldsymbol{x}}_{t\mid{t-1}} = \hat{\boldsymbol{x}}_{{t-1}\mid{t-1}} + \boldsymbol{B}_t\boldsymbol{u}_t
\end{equation}

\begin{equation}
\setstretch{1.5}
\boldsymbol{P}_{t\mid{t-1}} = \boldsymbol{P}_{{t-1}\mid{t-1}}+\boldsymbol{Q}_t
\end{equation}

\begin{figure}
\centering
{
\includegraphics[width=8cm]{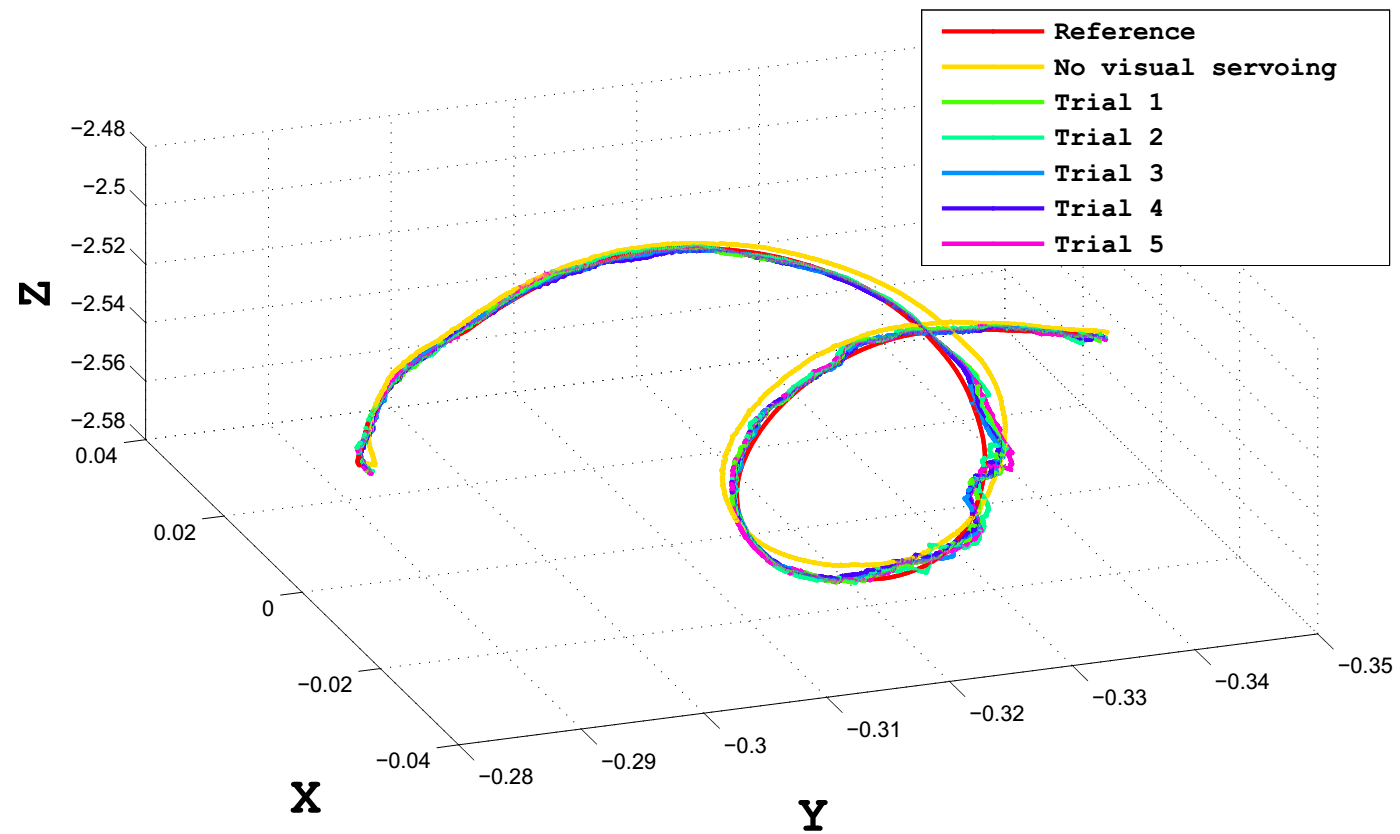}
\caption{Trajectory reproduction results demonstrating the relative errors for each trial compared to the reference.}
\vspace{-0.6cm}
\label{fig:opto_repeat}
}
\end{figure}

The symbol $\hat{\bullet}_{a\mid{b}}$ denotes the estimated value of $\bullet$ at time $a$ given its previous value at time $b$. The term $\boldsymbol{P}$ is the covariance matrix of the estimation of the state vector, and the term $\boldsymbol{Q}$ is, as mentioned above, the process noise covariance. At every time step, the robot was commanded to move to the next pose $x^*$ in the reference trajectory. Hence in our task the equation~\ref{equ:predict} is equivalent to

\begin{equation}
\hat{\boldsymbol{x}}_{t\mid{t-1}} = \boldsymbol{x}^*_{t-1}
\end{equation}

For the update phase, the current system state was estimated as

\begin{equation}
\hat{\boldsymbol{x}}_{t\mid{t}} = \hat{\boldsymbol{x}}_{t\mid{t-1}} +
\boldsymbol{K}_t\left(\hat{\boldsymbol{z}}_t-\hat{\boldsymbol{x}}_{t\mid{t-1}}\right)
\end{equation}
\begin{equation}
\boldsymbol{P}_{t\mid{t}} = \boldsymbol{P}_{{t}\mid{t-1}}+\boldsymbol{K}_t\boldsymbol{P}_{t\mid{t-1}}
\end{equation}
where the Kalman gain $\boldsymbol{K}_t$ is
\begin{equation}
\boldsymbol{K}_t = \boldsymbol{P}_{t\mid{t-1}}\left(\boldsymbol{P}_{t\mid{t-1}}+\boldsymbol{R}_t\right)^{-1}
\end{equation}

To compensate for camera latency, an additional prediction step was applied to estimate the current state measurement $\hat{\boldsymbol{z}}_t$. In our sewing task, this was the needle driver pose in the camera frame $\tensor[^c]{\hat{x}}{_d}$:
\begin{equation}
{
\hat{\boldsymbol{z}}_{t\mid{t_L}} =
\tensor[^c]{\hat{x}}{_d}\left(t\right) =
\tensor[^c]{{x}}{_d}\left({t}\right) \cdot
\left(\tensor[^r]x{_d}\left({t_L}\right)\right)^{-1} \cdot
\tensor[^r]x{_{d}}\left(t\right)
}
\end{equation}
where $t_L$ is the current time step minus the camera latency.

The filter runs in the same rate of the camera frame rate to provide visual feedback. Between the periods of two frames, robot follows the reference trajectory according to the last estimated needle driver pose.

The evaluation of the system state noise covariance $\boldsymbol{Q}$ and the measurement noise covariance $\boldsymbol{R}$ is explained in the next section.




\section{Experiments and results}
\label{sec:experiment}

\subsection {Estimation of Variables}

The covariance of the system noise $\boldsymbol{Q}$ and the measurement noise $\boldsymbol{R}$ were considered to be constant during the whole task. The system noise was dominated by the precision of the robot. According to the accuracy estimation~\cite{popic2015light}, the covariance matrix $\boldsymbol{Q}$ was estimated to be a diagonal matrix:

\begin{equation}
\small\boldsymbol{Q} = diag\left(0.25,0.25,0.25,0.02, 0.02, 0.02\right)
\end{equation}

An experiment was performed to evaluate the measurement error covariance matrix $\boldsymbol{R}$. In this experiment, the robot was moved slowly in a straight line while keeping the needle driver in the view of the camera. The needle driver pose was recorded from the camera and a straight line was fitted to the pose trajectory. According to the deviation of the detected poses to the straight line, the covariance matrix $\boldsymbol{R}$ was estimated to be a diagonal matrix:

\begin{equation}
\small\boldsymbol{R} = diag\left(0.35,0.35,0.35,0.04, 0.04,0.04\right)
\end{equation}

In both matrix, the first three values are in the unit of $mm$ and the later three are in $rad$.

\subsection {Task 1: Trajectory Reproduction}

\begin{figure}
\centering
{
\includegraphics[width=6cm]{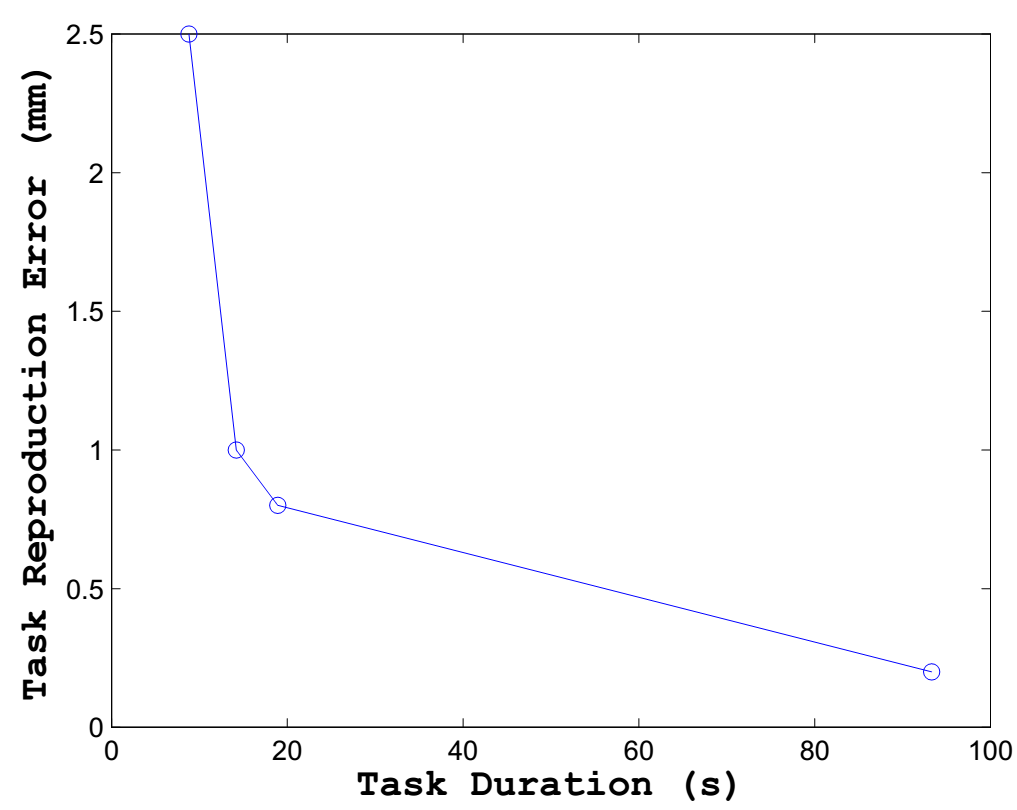}
\caption{Velocity vs. task duration for task reproduction.}
\vspace{-0.6cm}
\label{fig:Velocity-Time}}
\end{figure}


Detailed experiments with an Optotrak Certus (NDI) tracking device were conducted to validate the accuracy provided by visual servoing. A 3D printed rigid body (Fig.~\ref{fig:markers}) was designed to host both the infrared markers and our barcode markers. During each experimental trial, a ground truth trajectory was collected from user demonstrations and this trajectory was recorded both by the tracking device (infrared markers) and the camera (barcode markers). The robot was first commanded to execute the trajectory in two settings, with and without visual servoing, for which all actual movements were recorded by the tracking device. The actual movements resulted from both settings were then compared to the ground truth trajectory via dynamic time warping~\cite{Berndt1994} and rotation and translation differences between the trajectories were calculated. In this work, five trials were conducted and the results are provided in Fig.~\ref{fig:opto_repeat} and Table~\ref{tab:errors}. It can been seen from Fig.~\ref{fig:opto_repeat} that the trajectories executed without visual servoing deviate significantly from the ground truth for on average 5.23 mm per point while the trajectories generated with visual servoing are more accurate. Table~\ref{tab:errors} presents the quantitative results, from which we can observe that visual servoing improves the accuracy over no visual servoing guidance, with average error ranges of $\left[ 0.80,1.11\right]$ mm in translation and $\left[ 0.01,0.02 \right]$ degrees in rotation. Note that the errors presented in our visual servoing framework can result from marker tracking and pose estimation. This trajectory was reproduced with the same speed of demonstration. To analyse the correlation between the speed and the accuracy, another three sets of similar experiments were conducted with different speeds, i.e. reference trajectory sampling rate. The result is shown in Fig.~\ref{fig:Velocity-Time}. According to this result, we determine the sampling rate of the reference trajectory in different task contexts.

\begin{table}
\centering
\fontsize{8}{12}\selectfont
\caption{Trajectory reproduction accuracy}
\label{tab:errors}
\begin{tabular}{ccc}
\hline
\rowcolor[HTML]{C0C0C0}
Error              & Translation (mm) & Rotation (degree) \\ \hline \hline
No Visual Servoing & 5.23             & 0.07              \\ \hline
Trial 1            & 0.82             & 0.01              \\ \hline
Trial 2            & 0.81             & 0.01              \\ \hline
Trial 3            & 1.11             & 0.02              \\ \hline
Trial 4            & 0.92             & 0.01              \\ \hline
Trial 5            & 0.80             & 0.01              \\ \hline
\end{tabular}
\vspace{-0.2cm}
\end{table}

\begin{figure*}
\captionsetup[subfigure]{width=2.5cm}
\centering{
\includegraphics[width=18cm]{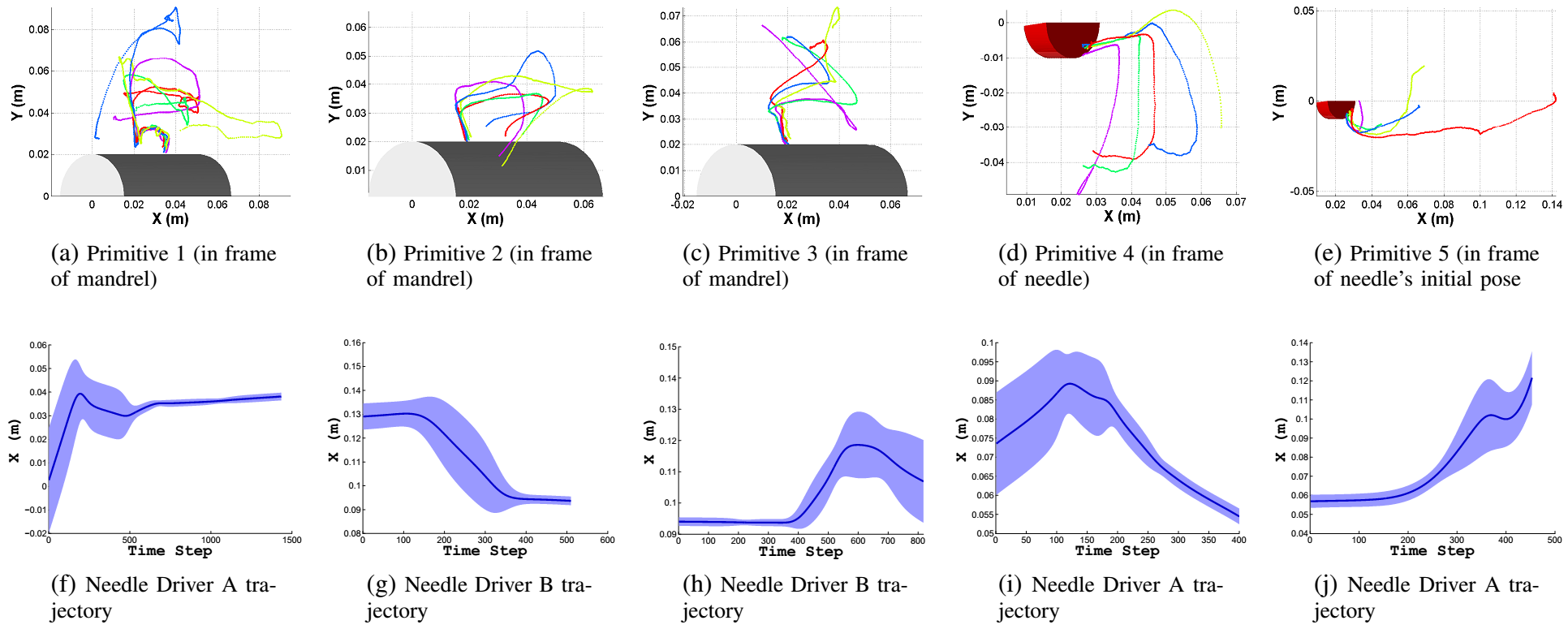}
\caption{The tip trajectories of the needle drivers from the user demonstrations and the generalized reference trajectories of all motion primitives in the bimanual sewing task. Top row: Multiple user demonstrations. Each colour represents one demonstration trajectory. (a)-(c): The grey cylinder represents the mandrel. (d)-(e): The red half cylinder represents the needle. Bottom row: 2D representation of the GMM of each primitive. Deep blue lines are the generalized trajectories, while light blue areas denote the corresponding variances.}
\label{fig:demo}
}\end{figure*}


\begin{figure*}
\centering
{
\includegraphics[width=18cm]{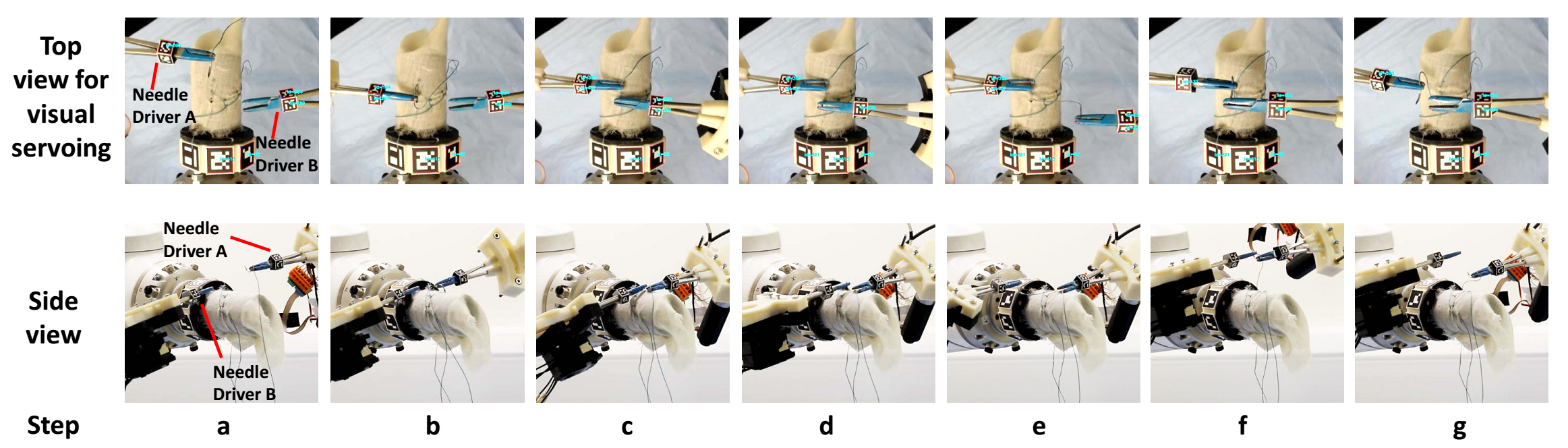}
\caption{Key frames of one stitch cycle of bimanual sewing. The top row shows the view from the top camera, which is used for visual servoing. The detected markers are labeled by red boxes and their IDs. The bottom row shows the corresponding views from the side. The letters at the bottom show their corresponding steps in Fig.~\ref{fig:stitchcycle}}
\vspace{-0.2cm}
\label{fig:bimanual}}
\end{figure*}

\subsection {Task 2: Robot Bimanual Sewing Task}

A bimanual sewing task was conducted to demonstrate the validity of this approach for teaching a robot a bimanual task. As mentioned in Section ~\ref{sec:demonstration}, in each stitching cycle the robot manipulates a needle to pierce the fabric bound on a mandrel. The stitching slots on the mandrel were of size of 2 by 10 $mm$ and hence high accuracy of reproducing the demonstrated trajectory was crucial. A small error in reproduction will cause an entire task failure.

Figure~\ref{fig:setup} shows the setup of our bimanual sewing system with three 7 d.o.f robots used (KUKA LWR 4+ and iiwa). To register all three robots to the camera frame, hand eye calibration was conducted as detailed in Section ~\ref{sec:demonstration}. The robot system was shown five cycles of stitching. A stereo vision system composed of two Logitech C920 web cameras retrieves 640$\times 480$-pixel videos for needle driver motion tracking and pose estimation. The trajectories of the needle drivers were registered to their corresponding sewing slots. The sewing slot positions were computed via the detected mandrel pose according to the mandrel design. A curved surgical curved needle of diameter 8$mm$ was used. Before each cycle (Figure~\ref{fig:stitchcycle}), the needle pose was detected to correct the needle driver trajectory so that the needle can be delivered to pierce accurately. At Step g, the robot was required to pull the thread and tighten the stitch; this pulling motion was performed manually according to an estimation of the remaining length of the thread. The mandrel is fixed on the third robot, of which the trajectory is programmed by mandrel's design, i.e. the sewing slots' locations.

Based on the variance of the demonstrations, the speed of the robot was varied in order to satisfy both the speed and accuracy requirements. Figure~\ref{fig:demo} shows the variance among all the demonstrations for Motion Primitives 1-5. It can be seen in Motion Primitive 1, both from the trajectories plot in the top row and the 2D representations of the GMM in the bottom row, that the motion of Needle Driver A has large variance at the beginning and small variance toward the end. Among all the demonstrations, the needle driver follows the same path to pierce the needle. For Motion Primitives 2 and 3, the motion of Needle Driver B has small variance when it is piercing out the needle (end of Motion Primitive 2 and beginning of Motion Primitive 3) and otherwise large variance. In Motion Primitive 4, Needle Driver A goes to pick up the needle from Needle Driver B. Hence the variance of Needle Driver A is large and then small in the frame of the needle. For the same reason, the variance of Needle Driver A in Motion Primitive 5 is small and then large. The motion of Needle Driver B in Motion Primitive 1, 4, 5 and of Needle Driver A in Motion Primitives 2 and 3 are small and hence considered to be static.

Overall, the variance is large for more than half of the task duration and therefore speeding up these parts of the task effectively speeds up the entire stitching cycle. The accuracy required in this task is preserved as the robot slows down at the moments requiring high precision. The robots performed 8 stitches with 6 success and 2 failures: one caused by the entanglement of the thread and one by the robot joint limit. The overall success rate of this bimanual sewing task is $75\%$.

\section{Discussion and Conclusion}

In this paper, we present a vision based programming by demonstration approach that enables users to demonstrate bimanual tasks with tools in their own hands and enables the robot to execute the task with varying speed according to the task context. Using the same tools and markers among the task demonstration and reproduction, the demonstrated motion skills can be transferred to robots seamlessly. The visually controlled execution allows for execution of a bimanual task with both robot position calibration and handeye calibration provided to a rough accuracy.

When the task context is end point driven, i.e. large deviation from the reference trajectory can be tolerated, the robot speeds up to reach the end point. When the task context is contact driven, i.e. the difference between the reproduced trajectory and the reference trajectory has to be minimised, the robot slows down. A trajectory reproduction task is conducted to evaluate the accuracy of visual servoing. The results show that as the robot speeds up, the accuracy decrease dramatically. According to this evaluation, we find out the speed corresponding to the required accuracy for the next sewing task.

In a typical bimanual sewing task, delicate skill and bimanual cooperations are involved such as piercing the needle in and out of the fabric via a narrow sewing slot and passing the needle over from one needle driver to another. The entire stitch cycle is segmented into five motion primitives and for each, the variance is computed. The robot reproduces the task with varying speed according to the corresponding variance. We show that the robot system can reproduce the original hand stitch with the guidance of a low cost camera. Despite the low sampling rate and camera latency, the robot system is able to reproduce the task with user level speed.

It should be noted that the thread has yet to be considered in our bimanual sewing task. To pull the thread, the needle currently moves out of the view of the stereo camera and we program this manually. The thread length was estimated as the original length minus the length of each stitch. The thread was pulled by moving the needle driver out of the the camera view from the bottom right and brought back into view from the top right. In this way, the thread formed a loop under its weight and a blanket stitch was formed, which is required for sewing a stent graft. To handle the thread properly in the future, a vision method is required to detect and localize the thread. As the current camera's view limits the working space for visual servoing, a multiple view system will be investigated.


Compared to kinesthetic teaching and tele-operation, our approach allows for a simple method to program multiple robots and benefits from allowing users to demonstrate bimanual tasks more naturally, demonstrations without the robot system, and the use of robots without built-in ``record-and-replay'' modes to be programmed.
We show that by combining the programming by demonstration and visual feedback control approaches, a low cost camera can supervise a multi-robot system to accomplish delicate tasks such as hand sewing. The proposed method in this paper can be applied to many other tasks involving cooperative control of multiple robots.




\bibliographystyle{IEEEtran}
\bibliography{IROS17}

\end{document}